\documentclass{article}

\PassOptionsToPackage{numbers, compress}{natbib}
\usepackage{genai4health_2026}

\usepackage[utf8]{inputenc} 
\usepackage[T1]{fontenc}    
\usepackage{hyperref}       
\usepackage{url}            
\usepackage{booktabs}       
\usepackage{amsfonts}       
\usepackage{nicefrac}       
\usepackage{microtype}      
\usepackage{xcolor}         
\usepackage{graphicx}       
\usepackage{caption}        
\usepackage{tikz}           
\usepackage{verbatim}
\usetikzlibrary{arrows.meta}

\title{Are We Grading Properly? Understanding Failure Modes in Medical Benchmarks}

\author{%
  \mdseries
  \begin{tabular}[t]{c}
    \textbf{Prithvi Dixit}$^{1,2}$\thanks{Corresponding author.} \\
    $^1$University of California, Berkeley \\
    \texttt{prithvi.dixit@berkeley.edu}
  \end{tabular}%
  \hspace{0.5em}%
  \begin{tabular}[t]{c}
    \textbf{Pedram Hosseini}$^2$ \\
    $^2$Medical Sphere \\
    \texttt{pedram@medicalsphere.ai}
  \end{tabular}
}

\begin{document}

\maketitle

\begin{abstract}
Medical evaluation is shifting from static option-based questioning to realistic clinical scenarios with open-ended output modes. Grading these at scale naively, however, is expensive, and rubric-based evaluation has become the dominant scalable alternative. We ask what happens when the rubrics themselves are not airtight, and whether such flaws can be detected and corrected. We apply RIFT \citep{qi2026riftrubricfailuremode}, a global rubric failure taxonomy, to three clinical benchmarks (HealthBench Professional \citep{healthbenchpro}, LiveMedBench \citep{yan2026livemedbenchcontaminationfreemedicalbenchmark}, and MentalHealthBench \citep{mentalhealthbench}), and find failure modes are meaningful: on HealthBench Professional an LLM judge flags 29.6\% of criteria as non-atomic and 65.4\% as misaligned/rigid, although these rates vary severalfold across five detection judges. Then, we show that these flaws are meaningful and not simply cosmetic. As an example, rewriting bundled criteria of the form ``at least one of / all of the following'' as equally weighted children and regrading identical responses shifts scores by up to 15.9 percentage points on affected conversations, with disjunctive bundles inflating scores and conjunctive bundles deflating them. Across six answering models and two common graders, these shifts leave the model ranking unchanged, and their net direction depends on how many bundled criteria are penalties. We also find that RIFT can under-detect bundling on clinical rubrics, by an amount that depends on the detection judge: on the same LiveMedBench criteria, judges flag 3.8--15.9\% as non-atomic against an estimated 7--19\% that are genuinely bundled.
\end{abstract}

\section{Introduction}
\label{sec:intro}

Medical evaluation is moving away from multiple choice to open-ended responses on real-world clinical scenarios. Although a variety of multiple choice benchmarks remain in use (MedQA, PubMedQA, MMLU-Med \citep{jin2020diseasedoespatienthave, jin2019pubmedqadatasetbiomedicalresearch, hendrycks2021measuringmassivemultitasklanguage}), evaluation for new models has shifted towards open-ended benchmarks \citep{healthbenchpro, yan2026livemedbenchcontaminationfreemedicalbenchmark}.

For open-ended benchmarks, cost becomes a dominant concern. Grading requires expert judgement, which is hard to scale given constraints on human labor and availability. Rubric-based grading with an LLM judge is the dominant answer, because physicians can write weighted criteria once and a model can apply them consistently at scale \citep{gu2025surveyllmasajudge, li2026llmasajudgehealthcarescopinganalysis}. Notably, nearly all measurement effort goes into model scores, while very little goes into whether the rubric itself is a valid instrument. A rubric is treated as ground truth by construction, but more stringent checks are needed before deployment. While a physician's authorship might guarantee clinical correctness, it does not guarantee measurement validity.

If the rubric itself is flawed, how do we detect it, and how does it change the numbers? Our hypothesis is that clinical rubrics contain systematic failure modes and that those failures propagate into reported model scores. We make the following contributions. First, we apply RIFT \citep{qi2026riftrubricfailuremode}, a global rubric failure taxonomy, to three clinical benchmarks and show that rubric failure modes are pervasive rather than occasional: on HealthBench Professional, our primary detection judge flags 29.6\% of criteria as non-atomic and 65.4\% as misaligned or rigid. These rates vary substantially across detection judges, so we also report the consensus of five. Second, we show these flaws are consequential rather than cosmetic: atomicizing bundled criteria and regrading identical responses shifts HBP scores by up to 15.9 percentage points on affected conversations, with disjunctive and conjunctive bundles moving scores in opposite directions as predicted; across six answering models and two common graders, these shifts do not reorder the models. Third, we replicate the direction of this effect on LiveMedBench at substantially smaller magnitude, observe it reverse on MentalHealthBench, and identify what makes a benchmark susceptible, including the share of bundled criteria that are penalties. Finally, we show that general-purpose RIFT can under-detect bundling in clinical rubrics, by an amount that depends strongly on the detection judge, so detection rates should be reported across several judges.

\section{Method}
\label{sec:method}

HealthBench Professional (HBP)\citep{healthbenchpro} and its rubric score a model's response to free-form questions against a set of criteria written and vetted by physicians. Every criterion is a requirement written that carries a point value, graded by an LLM judge that returns \texttt{criteria\_met} grading decisions using the  template released with HealthBench \citep{arora2025healthbenchevaluatinglargelanguage}; a conversation's score is the sum of points for met criteria, which then gets normed by the max available points. RIFT \citep{qi2026riftrubricfailuremode} is a general breakdown of ways that assumption can fail, labeling each unit with zero or more of eight modes: four are \emph{criterion-scope} (subjective, non-atomic, ungrounded, misaligned/rigid) and describe a defect in an individual scored item, and four are \emph{rubric-scope} (missing criteria, hackable, low-signal, redundant) and can only be assessed across a full rubric. Table~\ref{tab:rift} in the appendix gives the definitions.

\paragraph{Atomicization: split-and-regrade.}

Identifying this failure mode in LLM-written rubrics raises a follow-up question: do such instances have any impact on grading? To investigate, we introduce a \textbf{split-and-regrade} approach, illustrated in Figure~\ref{fig:method}. Say we have a criterion worth $P$ points that is written as a list of $N$ independent criteria. We turn this single criterion into $N$ atomic criteria each worth $P/N$ points. We do not change any other criteria in the rubric. Then we ask the \emph{same} grader to grade the \emph{same} responses both before and after splitting. Any change in grading is therefore attributable to the structure of the rubric. Moreover, we have a prediction of the direction of impact: a conjunctive criterion will lead to an increase in scores after splitting because partial credit is now possible; a disjunctive criterion will lead to a decrease in scores after splitting, as $P$ is awarded if \textit{at least} one subcriterion is met; after splitting, each criterion that is met awards $P/N$ points. These predictions allow us to control for artifacts that might affect grading before-and-after splitting, such as the grader becoming more/less lenient over time, which would lead to increases/decreases for conjunctive \textit{and} disjunctive criteria alike; our hypothesis on bundling is supported if we observe opposite-signed changes. These predictions hold for criteria that award points; for a penalty criterion they reverse, since a disjunctive penalty is charged in full when any one branch is met, so splitting it reduces the charge and raises the score.

\begin{figure}[t]
\centering
\begin{tikzpicture}[
  font=\footnotesize,
  parent/.style={draw, rounded corners=2pt, align=center, inner sep=4pt,
                 text width=32mm, fill=black!4},
  child/.style={draw, rounded corners=2pt, align=center, inner sep=2pt,
                text width=17mm},
  note/.style={align=left, inner sep=2pt, text width=34mm},
  arr/.style={-{Latex[length=1.8mm]}, semithick}
]

\node[parent] (p1) at (0,1.55) {\textbf{Disjunctive}\\``at least one of\\X, Y, Z''\quad $P{=}3$};
\node[child] (x1) at (4.15,2.17) {X\ \ $1$};
\node[child] (y1) at (4.15,1.55) {Y\ \ $1$};
\node[child] (z1) at (4.15,0.93) {Z\ \ $1$};
\draw[arr] (p1.east) -- node[above, midway, font=\scriptsize] {split} (3.32,1.55);
\node[note] (n1) at (7.35,1.55) {Meets X only:\\ $3/3 \;\rightarrow\; 1/3$\\ \textbf{score falls}};

\node[parent] (p2) at (0,-0.55) {\textbf{Conjunctive}\\``all of\\X, Y, Z''\quad $P{=}3$};
\node[child] (x2) at (4.15,0.07)  {X\ \ $1$};
\node[child] (y2) at (4.15,-0.55) {Y\ \ $1$};
\node[child] (z2) at (4.15,-1.17) {Z\ \ $1$};
\draw[arr] (p2.east) -- node[above, midway, font=\scriptsize] {split} (3.32,-0.55);
\node[note] (n2) at (7.35,-0.55) {Meets X, Y:\\ $0/3 \;\rightarrow\; 2/3$\\ \textbf{score rises}};

\draw[dashed, black!35] (-1.9,0.5) -- (9.2,0.5);
\end{tikzpicture}
\caption{Split-and-regrade. A bundled parent worth $P$ points is rewritten as
$N$ equally weighted children; the rest of the rubric, the model responses,
and the grading judge are held fixed. The two bundle types are predicted to
move scores in opposite directions, which distinguishes a rubric-structure
effect from a change in grader leniency.}
\label{fig:method}
\end{figure}

\section{Experimental setup}
\label{sec:experiments}

\subsection{Benchmarks and responses}
\label{sec:benchmarks}

\textbf{HBP} \citep{healthbenchpro} is our primary testbed: 525 conversations and 1,135 physician-written criteria. \textbf{LiveMedBench} \citep{yan2026livemedbenchcontaminationfreemedicalbenchmark} is a larger, contamination-controlled benchmark of 2,756 cases and 16,702 criteria, scored as the mean over cases of $\mathrm{clip}\!\left(\sum_i p_i \cdot m_i / P^{+}\right)$ for points $p_i$, met-decisions $m_i$, and total positive points $P^{+}$. Its official leaderboard judge is GPT-4.1. \textbf{MentalHealthBench} (MHB) \citep{mentalhealthbench} contains 1,215 mental-health conversations in eleven languages with 5,262 expert-authored criteria, 39\% of them penalties; like HBP it marks bundles with ``at least one of / all of the following'' (609 and 34 criteria). LiveMedBench and MHB are included to test if the results of HBP are a property of a single benchmark or more globally a result of rubric bundling failure.

Split-and-regrade requires a response that can be graded twice. On HBP we collect responses from Gemini 3.1 Flash-Lite (n=525) and GLM-5.2 (n=512 after excluding errors). To compare models under a common grader, we add DeepSeek V4.1 Flash, GLM-5.3 Flash, Gemma 4 31B, and Qwen3.8 27B (525 each) and a partial Kimi K2.6 set (314); on MHB we use these four (1,215 each). On LiveMedBench, we do the same extraction from Gemini 2.5 Flash and Gemini 3.5 Flash-Lite. 
\subsection{Detection protocols and judges}
\label{sec:protocols}

Detection uses the same labels as RIFT, with the same prompt and implementation as in that paper. For detection, we have two protocols. In the \textbf{Joined} protocol, we join all the criteria in a conversation into a single input to the detector, which must classify the conversation for all eight failure modes simultaneously. In the \textbf{Scoped} protocol, we call the detector once per criterion for the criterion-scope failure modes and once per conversation for the rubric-scope failure modes. On HBP, this corresponds to 1,135 items. While the Joined protocol may be advantageous if one cares about detection, we prefer the latter protocol as it provides a failure mode for each item, which is necessary to split-and-regrade. We primarily use GPT-5.4 as our detection judge, with one vote and the original, eight-mode RIFT prompt; see Section~\ref{app:figures} for further results using Gemini 3.1 Flash-Lite as the judge with one vote, GPT-5.4 mini as the judge with one vote, and GPT-5.4 mini as the judge with three-vote majority vote. 

For grading, we use the original \texttt{criteria\_met} prompt from the benchmarks. On HBP, we use Gemini 3.1 Flash-Lite and GPT-5.4 as our grading judge. To compare answer models fairly, we add two common graders, Qwen3.8 27B and GLM-5.3 Flash, each of which grades every HBP and MHB answer set (MHB with the HealthBench template). On LiveMedBench, the official grading judge is GPT-4.1, but to save costs, we use Gemini 3.5 Flash-Lite as our grading judge and regrade both answer sets with Qwen3.8 27B, meaning that our absolute scores are not comparable to those reported by the leaderboard. However, scores can be compared within our experimental setup. Confidence intervals are 95\% bootstrap intervals over conversations; rank stability is Kendall's $\tau$ between the original and atomicized leaderboards. Results are computed using unrounded averages and rounded only upon reporting.

\subsection{Identifying bundled criteria}
\label{sec:splitter}

HBP marks bundles explicitly with two recurring templates, ``at least one of the following'' (disjunctive) and ``all of the following'' (conjunctive), so parents and branches can be extracted directly from the criterion text. LiveMedBench does not have a similar setup. Therefore, we  created a rule-based splitter that identifies bundles from English surface forms and then tuned it over different recall attempts: it skips hedged constructions, temporal expressions, and nested lists because the cost of splitting a real requirement is too high. It identifies 4,310 bundled parents (2,826 disjunctive, 1,484 conjunctive) expanding to 9,989 children, covering 25.8\% of the benchmark's criteria (25.1\% of the 500-case sample), against 58.9\% that carry a raw surface conjunction or disjunction (Figure~\ref{fig:fig5} in the appendix). On MHB, whose templates list noun phrases after a colon, we split that list and reattach the shared stem to each child (2,121 children from 642 parents). We audit the splitters' precision in Appendix~\ref{app:audit}.

\section{Results}
\label{sec:results}

\subsection{Failure modes are pervasive, but measured rates depend on the judge}

Under the scoped protocol, GPT-5.4 flags a failure mode for a large fraction of HBP criteria: 65.4\% misaligned or rigid, 40.7\% subjective, 29.6\% non-atomic, and 23.3\% ungrounded. Rubric-scope rates are higher still, with 97.7\% of conversations missing a criterion implied by the prompt and 89.7\% containing low-signal items (Figure~\ref{fig:fig1} in the appendix). The rates when we take them absolutely vary substantially across detection judges (Figure~\ref{fig:fig8}a): across five judges, non-atomic ranges from 11.4\% to 29.6\%, hackable from 1.6\% to 41.9\%, and low-signal from 12.4\% to 89.7\%. Judges also disagree on which criteria fail (mean pairwise $\kappa$ of 0.08--0.30 per mode), even though the same models agree closely on the \texttt{criteria\_met} decisions used for grading ($\kappa=0.74$--$0.88$; Figure~\ref{fig:fig8}b). We therefore treat single-judge rates as within-judge comparisons rather than gold rates, and report the majority-of-five consensus as a less judge-dependent estimate: 55.9\% misaligned or rigid, 20.6\% ungrounded, 16.4\% subjective, and 11.3\% non-atomic, with 89.1\% of conversations missing a criterion.

\begin{figure}[t]
  \centering
  \includegraphics[width=\linewidth]{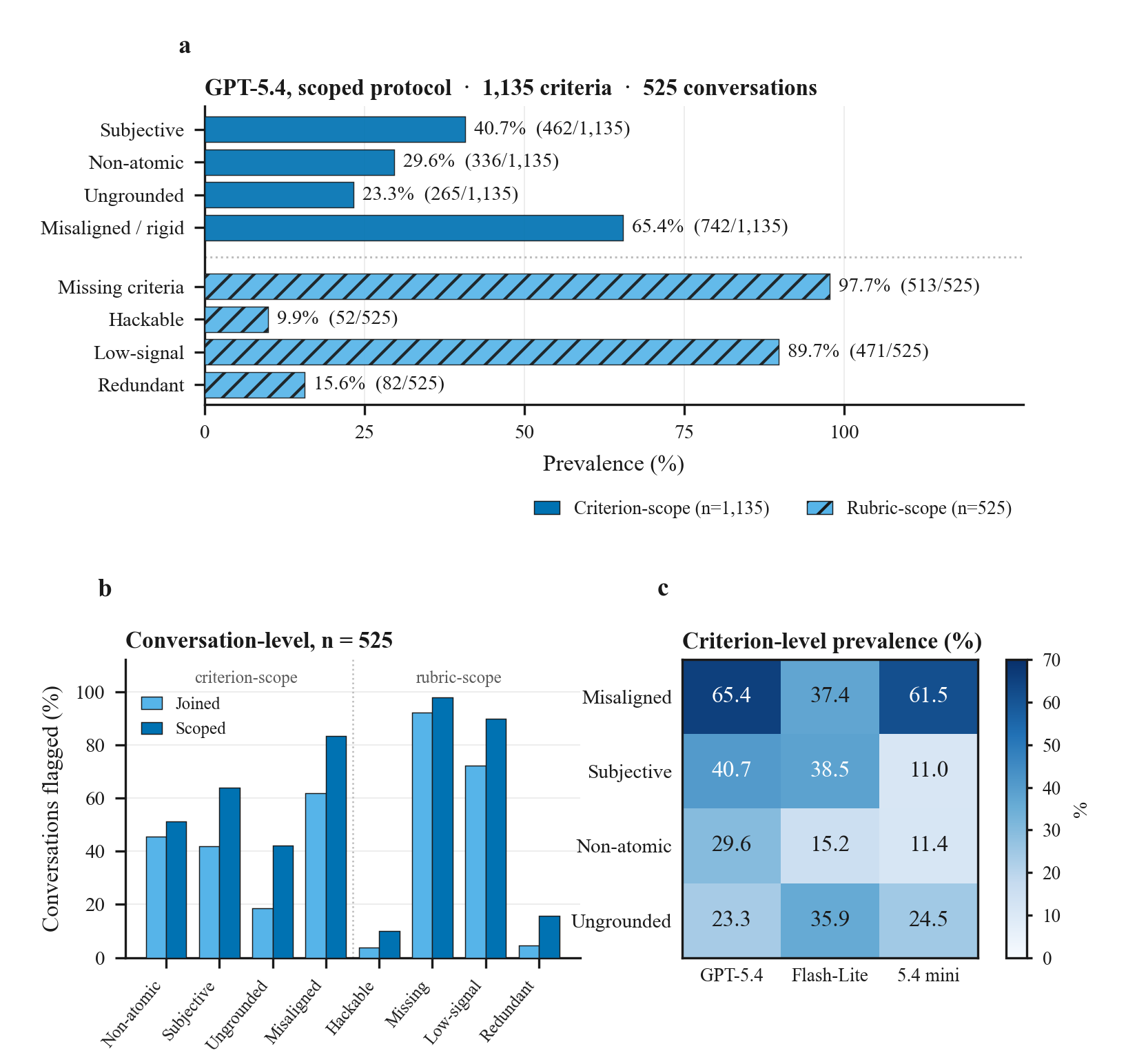}
  \caption{RIFT on HealthBench Professional. (a) GPT-5.4 scoped prevalence.
  Solid bars are criterion-scope (n=1,135); hashed bars are rubric-scope
  (n=525). (b) Conversation-level rates under joined (original RIFT protocol)
  vs.\ scoped. (c) Criterion-level prevalence by detection judge.}
  \label{fig:fig1}
\end{figure}

\subsection{Bundling changes reported scores}

\begin{figure}[t]
  \centering
  \includegraphics[width=0.85\linewidth]{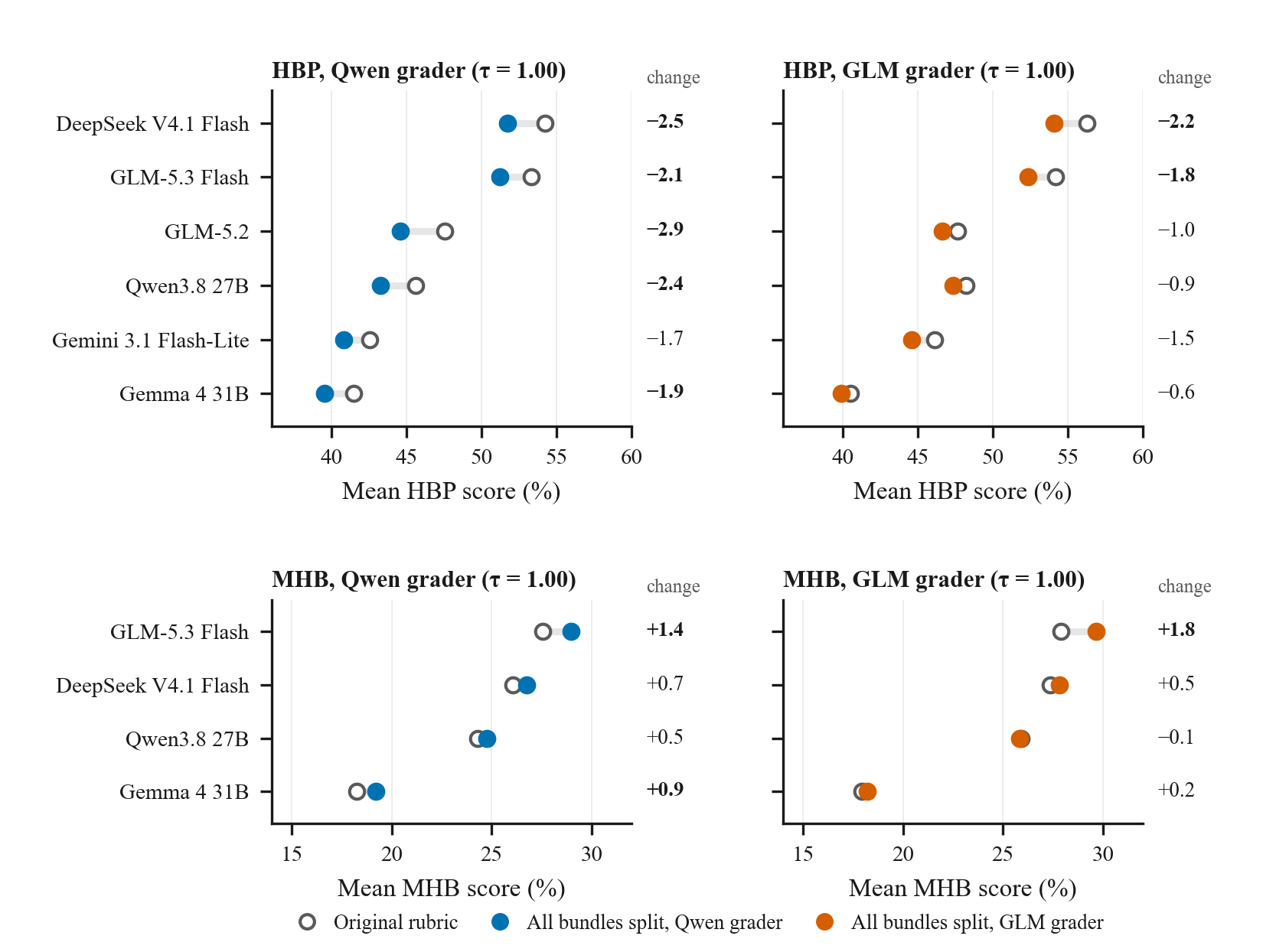}
  \caption{Atomicization shifts scores but not the ranking of models, on
  HealthBench Professional (top; six models, 512 shared conversations) and
  MentalHealthBench (bottom; four models, 1,215 conversations), each under two
  independent common graders. Hollow points are the original rubric, filled
  points the rubric with every bundled parent split; models are ordered by
  original Qwen-graded score. The number beside each row is the change in
  points, bold where its 95\% bootstrap interval excludes zero (intervals in
  Table~\ref{tab:leaderboard}). Splitting lowers HBP scores and raises MHB
  scores; Kendall $\tau=1.00$ in every panel.}
  \label{fig:fig6}
\end{figure}

Splitting the bundled criteria and regrading the same responses moves scores in every configuration we ran, and the two bundle types move them in opposite directions (Figure~\ref{fig:fig3} in the appendix). On the conversations that actually contain a split parent, disjunctive splits cost 8.4 points with Gemini 3.1 Flash-Lite as both answerer and grader, 8.2 points on GLM-5.2 responses, and 15.9 points when the same Flash-Lite responses are regraded by GPT-5.4. Conjunctive splits gain 18.9 points on their affected subset. Benchmark-wide effects are smaller ($-3.1$ to $-6.0$ points disjunctive, $+2.7$ conjunctive) simply because only a subset of conversations contains a bundled item.

The mechanism is visible at the branch level. Of the 219 disjunctive parents our branch parser could expand, the response satisfied exactly one branch in 52 cases: full credit under the original rubric, but $1/N$ of the item's points after the split. Because a grader-side shift in leniency would move both split types the same way, the sign reversal isolates the rubric's bundling structure as the cause rather than the judge.

These runs give each answer model its own grader. Under a common grader (Figure~\ref{fig:fig6}), on the 512 HBP conversations answered by all six models, disjunctive splits lower every model's score (by 4.0--4.5 points under Qwen3.8 27B and 2.4--3.8 under GLM-5.3 Flash) and conjunctive splits raise every model's score (by 1.3--2.7 points). With all bundles split, scores fall by 1.7--2.9 points under Qwen (five of six 95\% intervals exclude zero) and 0.6--2.2 under GLM, but the ranking of the six models is unchanged under both graders (Kendall $\tau=1.00$), and none of the 15 pairwise gaps changes significantly. Bundling therefore biases absolute scores by a few points without reordering models at this level of separation (Table~\ref{tab:leaderboard}).

The same procedure on LiveMedBench reproduces the direction but not the magnitude: disjunctive splits lower the mean clipped score by 0.43 and 0.58 points for the two answerers (0.18 and 0.27 when regraded by Qwen3.8 27B), conjunctive splits raise it by under 0.2 points (about 0.6 under Qwen), and effects on affected cases stay below one point (Figure~\ref{fig:fig4} in the appendix). Appendix~\ref{app:livemed} examines why the effect is small. We regard the benchmark-dependence as the substantive finding: the artifact is structural, but how much it moves a leaderboard depends on how much of the rubric is bundled, how the benchmark normalizes, and whether its bundles reward or penalize.

MHB shows the last factor directly. Atomicizing MHB \emph{raises} scores, by 0.45--1.42 points for all four answer models under the Qwen grader and by 0.24--1.75 points for three of the four under the GLM grader (the fourth is unchanged at $-0.08$), with rankings again unchanged ($\tau=1.00$; Figure~\ref{fig:fig6}, bottom). This is the penalty reversal predicted in Section~\ref{sec:method}: 53\% of MHB's disjunctive bundles are penalties (321/609), against 45\% on LiveMedBench and 11\% on HBP, and the net direction of the effect follows that share across the three benchmarks (Figure~\ref{fig:fig7}). Failure-mode rates on MHB are in Table~\ref{tab:mhbprev}.

Finally, the under-detection we report on LiveMedBench depends on the detection judge. Appendix~\ref{app:audit} compares the judges and the splitter on identical criteria.

\section{Conclusion}
\label{sec:conclusion}

Rubric-based grading has become the default way to evaluate language models on open-ended clinical tasks, and it carries an assumption that doesn't quite make sense today: that the rubric is a valid measuring instrument. We tested that assumption on three clinical benchmarks. Applying RIFT, we found failure modes in a large fraction of physician-written criteria, and by atomicizing bundled criteria and regrading identical model responses, we showed that at least one of those modes altered the scores reported in the direction its structure predicts, although it did not reorder the six models we compared. The effect has a consistent signature across benchmarks and graders, though its magnitude depends on how much of a rubric is bundled and its sign on whether bundles reward or penalize. While these results are promising, several limitations bound these results: our prevalence estimates come from LLM detection judges rather than human annotation, and absolute rates vary considerably across judges, so we treat them as within-judge comparisons rather than gold rates. Furthermore, our LiveMedBench grading runs use a lighter judge than the official leaderboard and is thus likely not comparable to published values; our claims rest on paired differences under a fixed grader, which hold under the stronger Qwen3.8 27B regrade. A score change under atomicization also shows that a rubric's aggregation rule matters, not that the original rule was wrong: ``at least one of'' or ``all of'' may be exactly what the author intended, and deciding which scoring better reflects response quality requires clinician judgments, which we leave to future work along with human labels of failure modes and of splitter output (we release a sampled annotation sheet). Finally, we tested only one failure mode, bundling, causally; the other seven are described, not shown to change scores.

We believe the practical implication is actionable today. Benchmark authors should write bundled requirements as separate scored items with points distributed across branches, because the alternative rewards partial answers in one direction and hurts in the other. More generally, the implication is that rubric quality should get the same measurement scrutiny that is given to the models rubrics are used to measure.


\bibliographystyle{unsrtnat}
\bibliography{citations}

\clearpage
\appendix
\raggedbottom

\section{Taxonomy definitions}
\label{app:taxonomy}

\begin{center}
  \captionof{table}{RIFT failure modes. Criterion-scope modes are applied to
  individual rubric criteria; rubric-scope modes are applied to the rubric as
  a whole.}
  \label{tab:rift}
  \small
  \begin{tabular}{llp{6.2cm}}
    \toprule
    Mode & Scope & Definition \\
    \midrule
    Subjective        & Criterion & Uses inherently subjective evaluative terms without sufficiently anchoring them with objective expectations. \\
    Non-Atomic        & Criterion & Bundles multiple independently scorable requirements into one item, preventing consistent partial credit or a parseable scoring structure. \\
    Ungrounded        & Criterion & Requires verification of a groundable or factually checkable claim without sufficient grounding or bounds, such as reference facts, sources, time scope, or a verification procedure. \\
    Misaligned/Rigid  & Criterion & Grades the wrong objective, embeds incorrect assumptions, or imposes unnecessarily strict or narrow requirements not asked for by the prompt. \\
    \midrule
    Missing Criteria  & Rubric & Omits a checkable requirement implied by the prompt, leaving no criterion through which a grader can evaluate it. \\
    Hackable          & Rubric & Can be gamed at the rubric level, allowing a responder to achieve a high score by inflating proxy metrics without materially improving correctness or fulfillment of the prompt. \\
    Low Signal        & Rubric & Fails to discriminate candidate responses well, assigning nearly equivalent scores to responses with substantively different quality. \\
    Redundant         & Rubric & Contains multiple criteria that evaluate the same underlying requirement, causing the same behavior to be rewarded or penalized more than once. \\
    \bottomrule
  \end{tabular}
\end{center}

\section{Common-grader leaderboards}
\label{app:leaderboard}

\begin{center}
  \captionof{table}{Mean score (\%) under the original rubric and with all bundled parents split, for every answer model under two common graders, on the conversations all models in a benchmark answered. $\Delta$ is the change in points with a 95\% bootstrap interval over conversations. Rankings are identical before and after splitting in every column (Kendall $\tau=1.00$). In a seven-model HBP view adding Kimi K2.6 on its 313 shared conversations, $\tau=1.00$ under Qwen and $0.90$ under GLM, where DeepSeek V4.1 Flash and GLM-5.3 Flash (0.3 points apart) swap.}
  \label{tab:leaderboard}
  \small
  \resizebox{\linewidth}{!}{%
  \begin{tabular}{lrrlrrl}
    \toprule
    & \multicolumn{3}{c}{Qwen3.8 27B grader} & \multicolumn{3}{c}{GLM-5.3 Flash grader} \\
    \cmidrule(lr){2-4}\cmidrule(lr){5-7}
    Answer model & Orig. & Split & $\Delta$ [95\% CI] & Orig. & Split & $\Delta$ [95\% CI] \\
    \midrule
    \multicolumn{7}{l}{\textit{HealthBench Professional} (6 models, 512 shared conversations)} \\
    \midrule
    DeepSeek V4.1 Flash    & 54.2 & 51.7 & $-2.5$ [$-4.1$, $-0.8$] & 56.3 & 54.1 & $-2.2$ [$-3.9$, $-0.5$] \\
    GLM-5.3 Flash          & 53.3 & 51.2 & $-2.1$ [$-3.9$, $-0.2$] & 54.2 & 52.3 & $-1.8$ [$-3.6$, $-0.0$] \\
    GLM-5.2                & 47.5 & 44.6 & $-2.9$ [$-4.6$, $-1.4$] & 47.6 & 46.6 & $-1.0$ [$-2.4$, $+0.5$] \\
    Qwen3.8 27B            & 45.6 & 43.2 & $-2.4$ [$-4.1$, $-0.6$] & 48.2 & 47.3 & $-0.9$ [$-2.7$, $+0.9$] \\
    Gemini 3.1 Flash-Lite  & 42.5 & 40.8 & $-1.7$ [$-3.5$, $+0.0$] & 46.1 & 44.6 & $-1.5$ [$-3.4$, $+0.2$] \\
    Gemma 4 31B            & 41.5 & 39.6 & $-1.9$ [$-3.7$, $-0.2$] & 40.5 & 39.9 & $-0.6$ [$-2.5$, $+1.2$] \\
    \midrule
    \multicolumn{7}{l}{\textit{MentalHealthBench} (4 models, 1,215 shared conversations)} \\
    \midrule
    GLM-5.3 Flash          & 27.6 & 29.0 & $+1.4$ [$+0.6$, $+2.3$] & 27.9 & 29.6 & $+1.8$ [$+0.8$, $+2.7$] \\
    DeepSeek V4.1 Flash    & 26.1 & 26.8 & $+0.7$ [$-0.2$, $+1.5$] & 27.4 & 27.8 & $+0.5$ [$-0.4$, $+1.4$] \\
    Qwen3.8 27B            & 24.3 & 24.7 & $+0.5$ [$-0.3$, $+1.2$] & 25.9 & 25.8 & $-0.1$ [$-0.8$, $+0.6$] \\
    Gemma 4 31B            & 18.3 & 19.2 & $+0.9$ [$+0.2$, $+1.7$] & 18.0 & 18.2 & $+0.2$ [$-0.4$, $+0.9$] \\
    \bottomrule
  \end{tabular}}
\end{center}

\vspace{1em}

\begin{center}
  \captionof{table}{RIFT failure-mode prevalence (\%) on MentalHealthBench.
  GLM-5.3 Flash labeled the full benchmark (5,262 criteria, 1,215
  conversations); Qwen3.8 27B labeled a 200-conversation subsample (880
  criteria), where both judges are compared. ``Both'' is the share flagged by
  both judges; $\kappa$ is their agreement on the subsample. Although 12\% of
  MHB criteria use explicit ``at least one of / all of the following''
  templates, both judges flag about 6\% of those templated items as non-atomic
  (GLM 42/643, Qwen 6/94), against 26\% of HBP's templated items for GLM
  (96/372).}
  \label{tab:mhbprev}
  \small
  \begin{tabular}{llrrrrr}
    \toprule
    & & Full benchmark & \multicolumn{4}{c}{200-conversation subsample} \\
    \cmidrule(lr){3-3}\cmidrule(lr){4-7}
    Mode & Scope & GLM & GLM & Qwen & Both & $\kappa$ \\
    \midrule
    Subjective        & Criterion & 10.9 & 11.7 & 10.5 & 5.6  & 0.44 \\
    Non-atomic        & Criterion & 3.6  & 4.0  & 1.7  & 1.1  & 0.39 \\
    Ungrounded        & Criterion & 11.5 & 12.3 & 13.3 & 3.5  & 0.17 \\
    Misaligned/rigid  & Criterion & 43.9 & 45.1 & 59.3 & 34.2 & 0.29 \\
    Hackable          & Criterion & 8.2  & 8.1  & 11.0 & 2.3  & 0.16 \\
    \midrule
    Missing criteria  & Rubric    & 64.0 & 65.5 & 60.0 & 49.0 & 0.41 \\
    Low-signal        & Rubric    & 28.7 & 26.5 & 22.5 & 12.0 & 0.33 \\
    Redundant         & Rubric    & 8.8  & 8.5  & 3.5  & 2.5  & 0.39 \\
    \bottomrule
  \end{tabular}
\end{center}

\section{Why the LiveMedBench effect is small}
\label{app:livemed}

Three candidate factors could explain the gap. Branch-satisfaction patterns differ sharply: only 9.5--11.4\% of LiveMedBench disjunctive parents were satisfied by exactly one branch, against 23.7\% on HBP, so far fewer items are positioned to change value. Our surface-form splitter is also stricter than HBP's explicit templates, but the lightweight grading judge is not the explanation: regrading with Qwen3.8 27B lowers absolute scores by about 6 points yet keeps every effect below one point (Figure~\ref{fig:fig7}).

\section{Splitter audit and under-detection}
\label{app:audit}

Because the splitter defines what counts as a bundle, we audited it: for each benchmark we sampled 150 split parents and asked two LLM auditors (DeepSeek V4.1 Flash and GLM-5.3 Flash) whether the parent bundles independently scorable requirements and whether its children faithfully restate it. Template-based splits score well (HBP 74\% and 91\%; MHB 91\% for both auditors), but the LiveMedBench splitter far less consistently (27\% and 73\%; inter-auditor $\kappa=0.34$; Figure~\ref{fig:fig9}b).

The splitter is not ground truth: on 903 of the sampled criteria it splits 26.0\%, which at its audited precision of 27--73\% corresponds to roughly 7--19\% genuine bundles. On the same criteria, RIFT flags 3.8\% as non-atomic under Gemini 3.1 Flash-Lite, 5.6\% under GLM-5.3 Flash, and 15.9\% under Qwen3.8 27B (Figure~\ref{fig:fig9}a). A weak judge misses most bundles; a stronger one flags roughly as many as the precision-adjusted splitter finds.

\section{Additional figures}
\label{app:figures}

\begin{center}
  \includegraphics[width=0.85\linewidth]{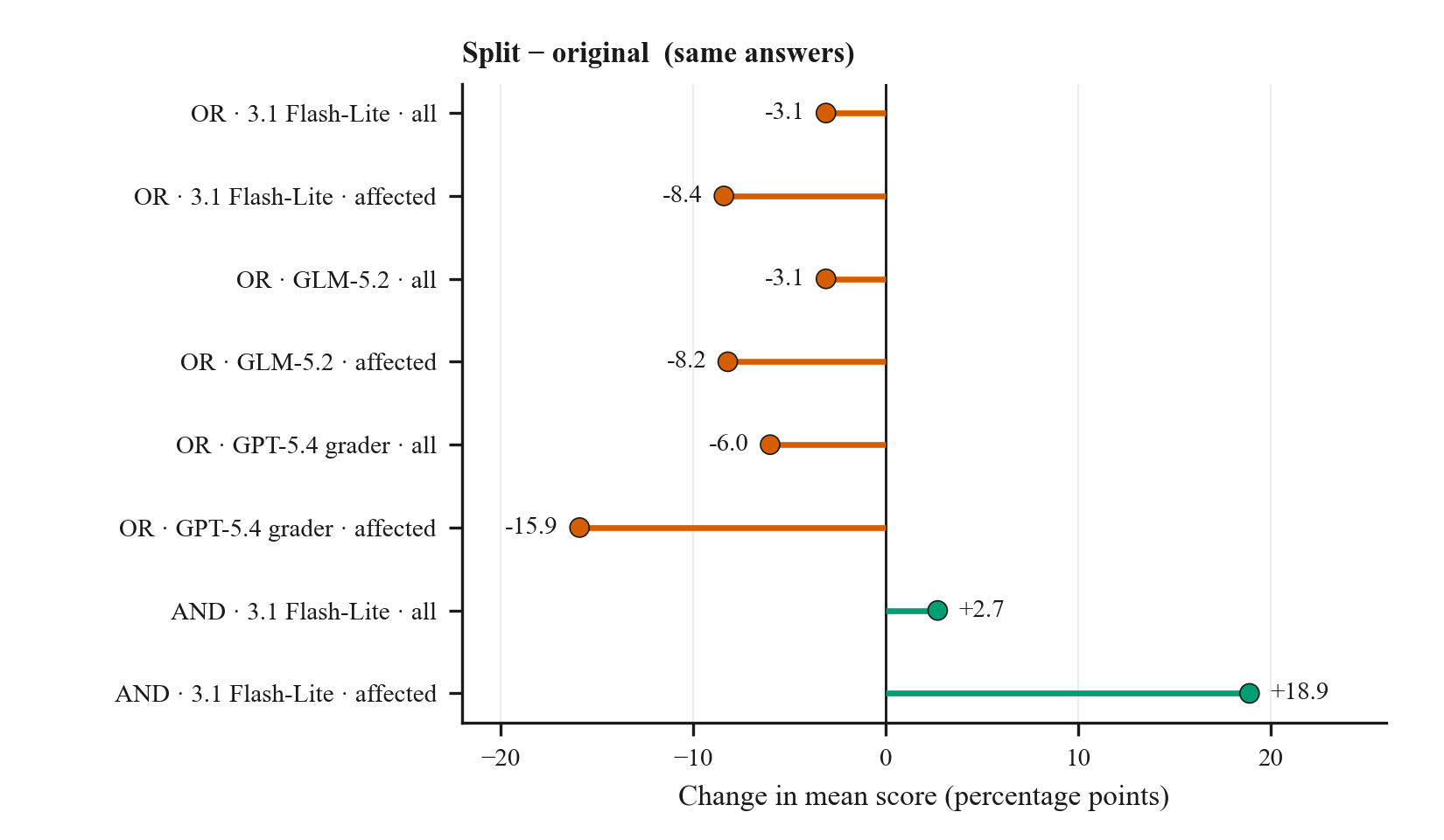}
  \captionof{figure}{Split minus original mean score on HBP (percentage points), same
  responses and same grading judge in every row. ``All'' is every conversation
  in the run; ``affected'' is the subset containing at least one split parent.
  Disjunctive and conjunctive splits move scores in opposite directions.}
  \label{fig:fig3}
\end{center}

\vspace{1em}

\begin{center}
  \includegraphics[width=\linewidth]{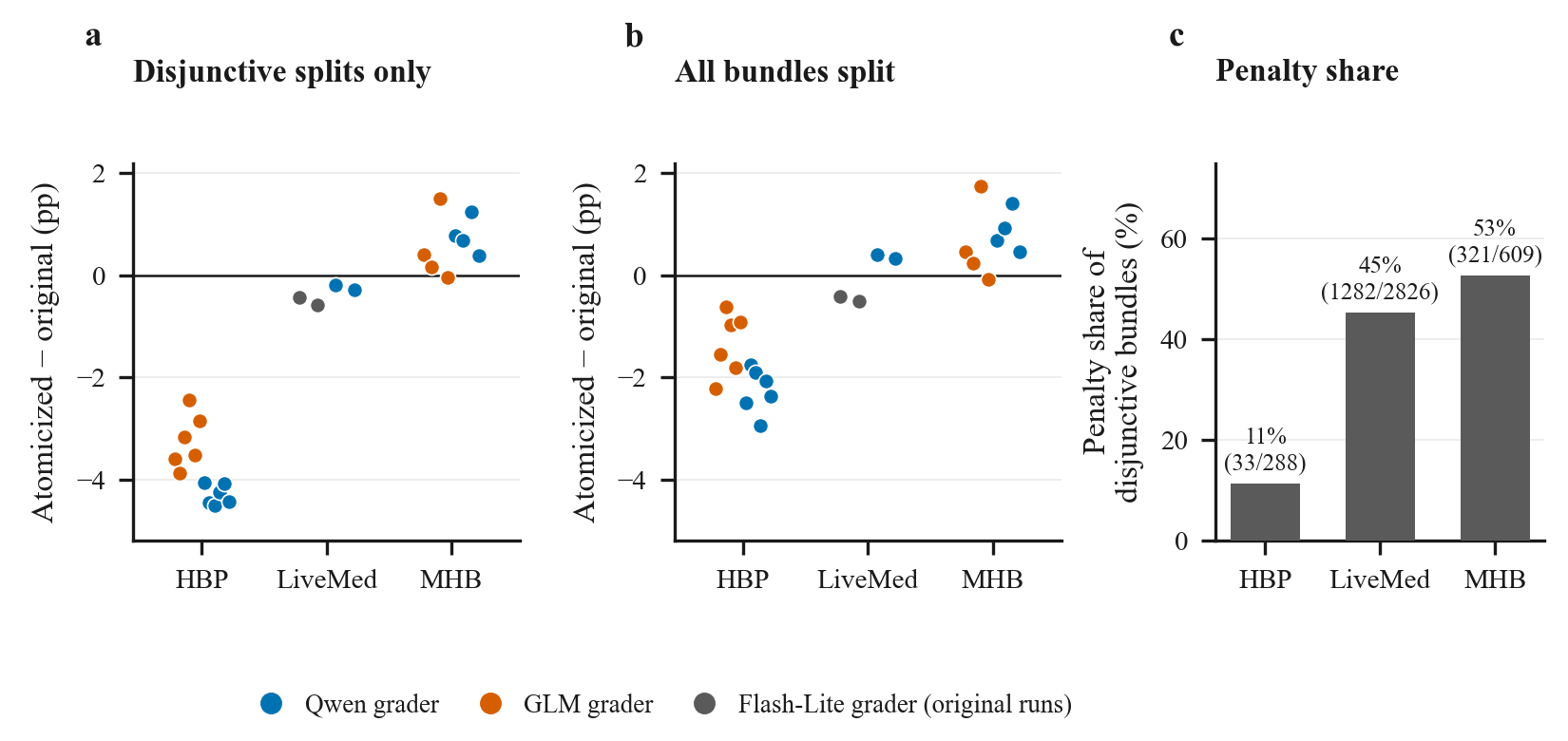}
  \captionof{figure}{The direction of the bundling effect depends on the benchmark. Each
  point is one answer model under one grader; benchmark-wide change in mean
  score (percentage points). (a) Disjunctive parents split. (b) All bundled
  parents split. (c) Share of disjunctive bundles that are penalties. HBP: six
  models, two common graders; MHB: four models, two common graders;
  LiveMedBench: two models, each under Gemini 3.5 Flash-Lite (gray, original
  runs) and Qwen3.8 27B. Blue = Qwen grader, vermilion = GLM grader.}
  \label{fig:fig7}
\end{center}

\vspace{1em}

\begin{center}
  \includegraphics[width=0.95\linewidth]{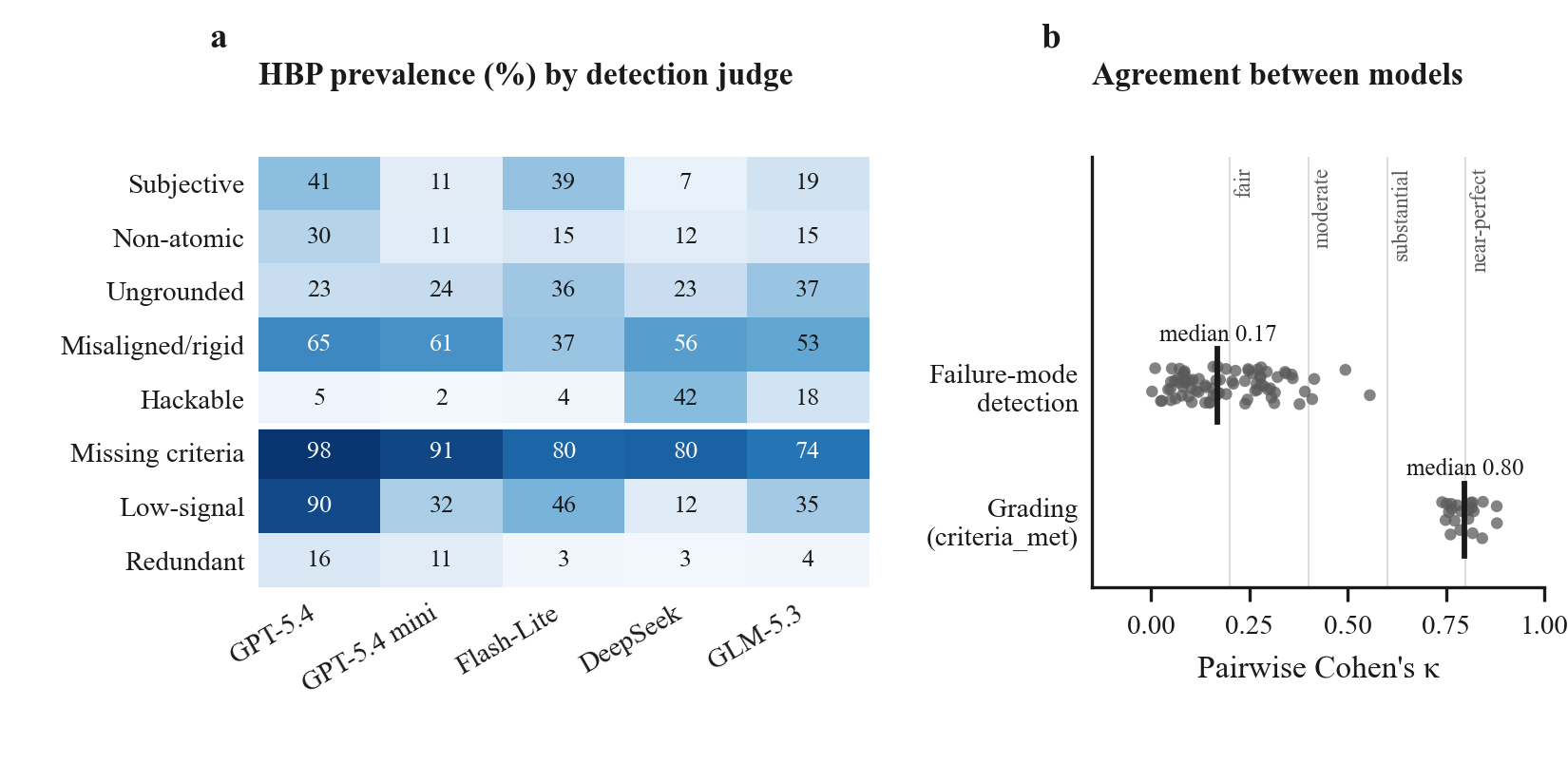}
  \captionof{figure}{Failure-mode detection is judge-dependent; grading is
  not. (a) Prevalence (\%) of each RIFT mode on HBP under five detection
  judges (criterion-scope modes over 1,135 criteria, rubric-scope modes over
  525 conversations). (b) Pairwise Cohen's $\kappa$ between models: each point
  in the upper row is one judge pair on one failure mode (80 points), each
  point in the lower row one pair of graders on item-level
  \texttt{criteria\_met} decisions for the same responses (22 points). Black
  bars mark medians; vertical lines mark conventional agreement bands.}
  \label{fig:fig8}
\end{center}

\vspace{1em}

\begin{center}
  \includegraphics[width=0.95\linewidth]{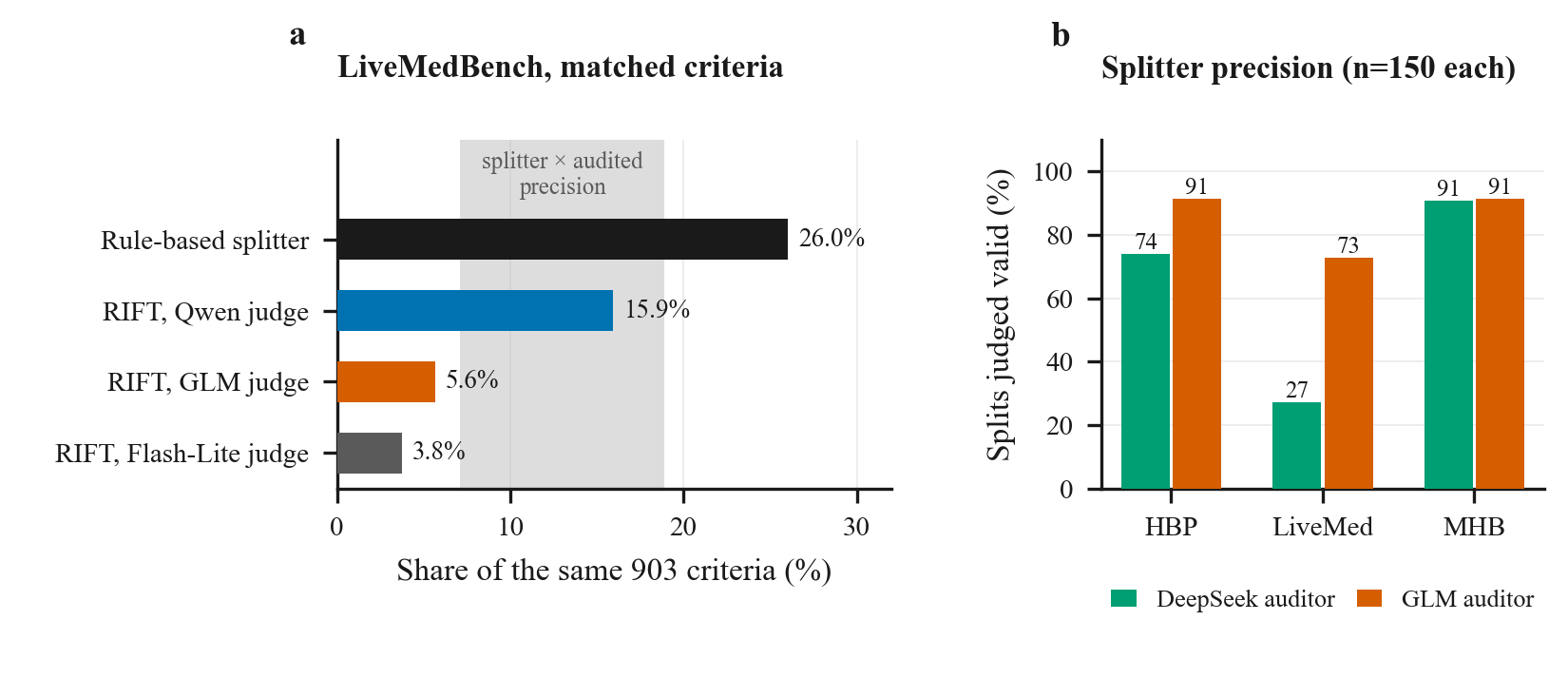}
  \captionof{figure}{Bundling under-detection on a matched denominator, and
  splitter precision. (a) Share of the same 903 LiveMedBench criteria (the
  first 150 of the 500 sampled cases) flagged as non-atomic by RIFT under
  three judges, and split by the rule-based splitter; the gray band is the
  splitter rate multiplied by its audited precision range. (b) Share of 150
  sampled split parents per benchmark judged to be genuine bundles with
  faithful children, by two LLM auditors (inter-auditor $\kappa=0.34$).}
  \label{fig:fig9}
\end{center}

\vspace{1em}

\begin{center}
  \includegraphics[width=0.95\linewidth]{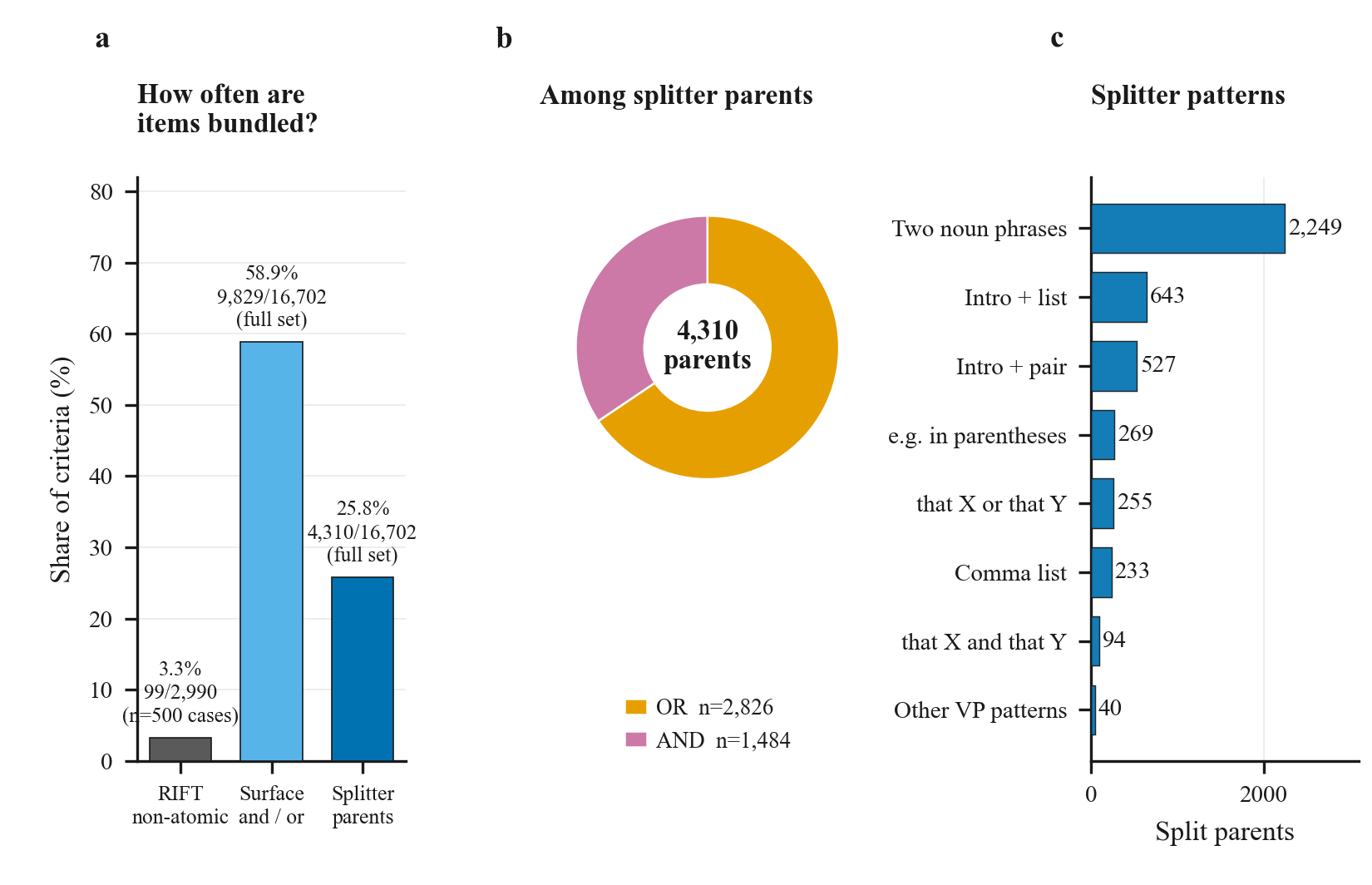}
  \captionof{figure}{Bundling prevalence on LiveMedBench. (a) Share of
  criteria identified as bundled by three methods: RIFT's non-atomic mode on a
  500-case sample (3.3\%, 99/2,990), raw surface conjunctions or disjunctions
  across the full set (58.9\%, 9,829/16,702), and our precision-oriented
  splitter (25.8\%, 4,310/16,702). Denominators differ: the RIFT rate is over
  the sampled subset, the other two over all 16,702 criteria. Figure~\ref{fig:fig9}a compares judges and the splitter on identical criteria.
  (b) Disjunctive/conjunctive composition of splitter parents. (c) Surface
  patterns matched by the splitter.}
  \label{fig:fig5}
\end{center}

\vspace{1em}

\begin{center}
  \includegraphics[width=0.95\linewidth]{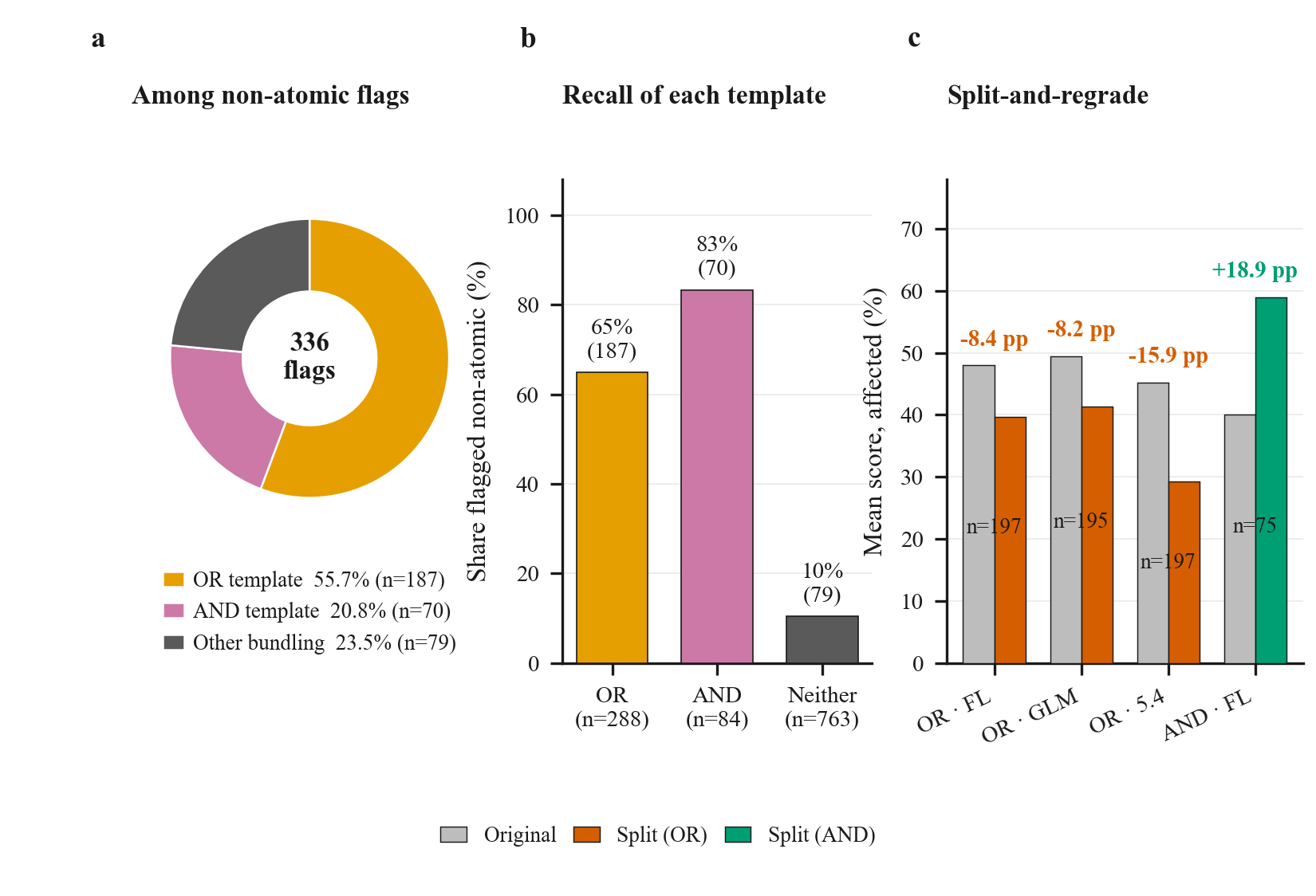}
  \captionof{figure}{Non-atomic bundling is mostly disjunctive or conjunctive,
  and splitting it changes scores. (a) Composition of GPT-5.4 non-atomic flags
  (n=336). (b) Fraction of HBP items with each surface form that RIFT flags as
  non-atomic. (c) Mean score on conversations that contain a split parent,
  original rubric vs.\ atomicized children, same answers. Disjunctive splits
  lower scores; conjunctive splits raise them.}
  \label{fig:fig2}
\end{center}

\vspace{1em}

\begin{center}
  \includegraphics[width=0.95\linewidth]{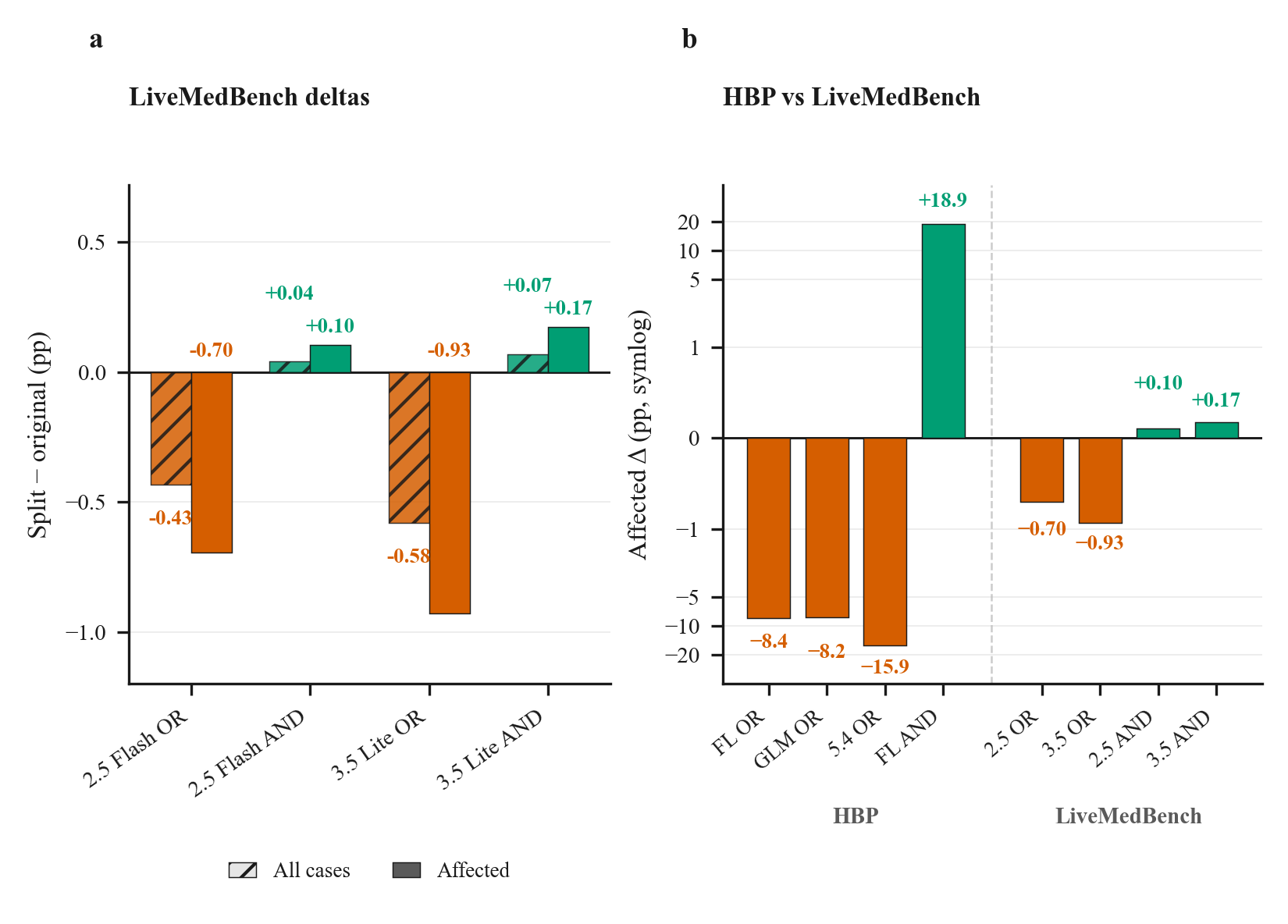}
  \captionof{figure}{Split-and-regrade on LiveMedBench. (a) Change in mean
  clipped score after atomicization, in percentage points; hatched bars are all
  cases, solid bars the affected subset (disjunctive n=1,718; conjunctive
  n=1,091). All runs use Gemini 3.5 Flash-Lite as grading judge (the Qwen3.8 27B regrade is in Figure~\ref{fig:fig7}).
  (b) Affected-case magnitude on HBP versus LiveMedBench, symlog scale.
  FL = Gemini 3.1 Flash-Lite responses, GLM = GLM-5.2, 5.4 = regraded by
  GPT-5.4. The direction replicates across benchmarks; the magnitude differs
  by more than an order of magnitude.}
  \label{fig:fig4}
\end{center}

\vspace{1em}

\begin{center}
  \includegraphics[width=0.62\linewidth]{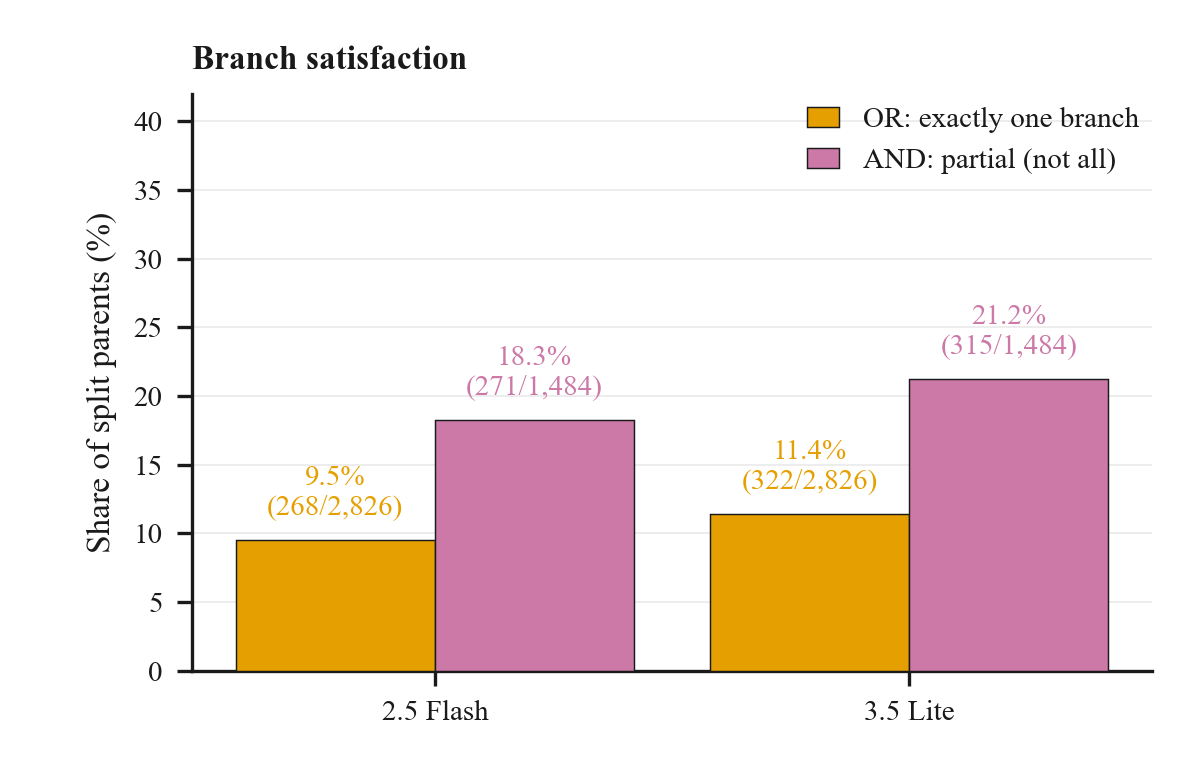}
  \captionof{figure}{Branch-satisfaction composition of LiveMedBench split
  parents. Disjunctive bars show the share of the 2,826 disjunctive parents
  that the response satisfied by exactly one branch; conjunctive bars show the
  share of the 1,484 conjunctive parents that were partially satisfied. Both
  are the configurations that change value under atomicization.}
  \label{fig:app-branch}
\end{center}

\end{document}